\documentclass[11pt]{article}
\usepackage[utf8]{inputenc}

\usepackage[left=23mm,right=23mm,top=23mm,bottom=23mm,paper=a4paper]{geometry}

\usepackage{graphicx} 
\usepackage{amssymb, amsmath, amsfonts, amsthm, graphics}
\usepackage{subfig}
\usepackage{multicol}
\usepackage{xcolor}
\usepackage[colorlinks = true,
            linkcolor = blue,
            urlcolor  = blue,
            citecolor = blue,
            anchorcolor = blue]{hyperref}
\usepackage{cleveref}
\usepackage{wrapfig}
\usepackage[square,numbers,sort&compress]{natbib}
\usepackage[font=small,labelfont=bf]{caption}
\usepackage{booktabs}

\usepackage{algorithm}
\usepackage[noend]{algorithmic}

\theoremstyle{plain}
\newtheorem{theorem}{Theorem}
\newtheorem{lemma}[theorem]{Lemma}
\newtheorem{fact}[theorem]{Fact}

\newcommand{\eat}[1]{}

\newcommand{\RR}{\mathbb{R}}
\renewcommand{\SS}{\mathbb{S}}

\newcommand{\cA}{\mathcal{A}}

\newcommand{\cX}{\mathcal{X}}
\newcommand{\cY}{\mathcal{Y}}

\DeclareMathOperator*{\argmin}{arg\,min}
\DeclareMathOperator*{\minimize}{minimize}
\DeclareMathOperator*{\maximize}{maximize}
\DeclareMathOperator*{\st}{subject\;to}

\newcommand{\adj}{\ensuremath{\mathclose{\vphantom{)}}^*}}

\newcommand{\tr}{\mathrm{tr}}

\title{Polynomial Precision Dependence Solutions to\\ Alignment Research Center Matrix Completion Problems}
\author{Rico Angell\thanks{Manning College of Information and Computer Sciences, University of Massachusetts Amherst, Amherst, MA, USA. Correspondence to Rico Angell (\href{mailto:rangell@cs.umass.edu}{rangell@cs.umass.edu}).}}

\begin{document}

\maketitle

We present solutions to the matrix completion problems proposed by the Alignment Research Center\footnote{\url{https://www.alignment.org/blog/prize-for-matrix-completion-problems/}} that have a polynomial dependence on the precision $\varepsilon$.
The motivation for these problems is to enable efficient computation of heuristic estimators to formally evaluate and reason about different quantities of deep neural networks in the interest of AI alignment~\cite{christiano2022formalizing}.
Our solutions involve reframing the matrix completion problems as a semidefinite program (SDP) and using recent advances in spectral bundle methods for fast, efficient, and scalable SDP solving.
The two matrix completion problems are restated as follows:

\paragraph{Question 1 (existence of positive semidefinite completions):} \!\!Given $m$ entries of a symmetric matrix $M \in \SS^n$, including the diagonal, can we tell in time $\tilde{O}(nm)$ whether it has any (real, symmetric) positive semidefinite completion? Proving that this task is at least as hard as dense matrix multiplication or positive semidefinite testing would count as a resolution.

\paragraph{Question 2 (fast ``approximate squaring''):} \!\!Given $A \in \RR^{n \times n}$ and a set of $m = \Omega(n)$ entries of $A A^T =: M \in \SS^n$, can I find \emph{some} positive semidefinte matrix that agrees with $M$ in those $m$ entries in $\tilde{O}(n m)$ time? \\

Instead of achieving a logarithmic dependence in the precision $\varepsilon$ (suppressed by $\tilde{O}(\cdot)$ in the original questions)\footnote{ It was suggested in a later blog post (\url{https://www.alignment.org/blog/matrix-completion-prize-results/}) by the Alignment Research Center that it is unlikely to be able to solve these problems without either at least $\Omega(n^3)$ time or a polynomial dependence on $\varepsilon$. Our results somewhat support this statement as will be seen.}, we present a solution to Question 1 in $\tilde{O}(\min\{ nm / \varepsilon, m / \varepsilon^{3/2}\})$ time and solutions to Question 2 in $\tilde{O}(n^2 / \varepsilon^2 + \min\{ nm / \varepsilon^2, m / \varepsilon^{5/2}\})$ and $\tilde{O}(n\sqrt{m}/\varepsilon^2 + \min\{ nm / \varepsilon^2, m / \varepsilon^{5/2}\})$ time, depending on whether matrix sketching is used or not.
The core idea is to formulate both questions as semidefinite programs (SDP) and use a spectral bundle method~\cite{helmberg2000spectral, helmberg2002spectral, helmberg2014spectral, ding2023revisiting, angell2023fast} to \emph{implicitly} solve the SDP or obtain a certificate of infeasibility.
In the case where the SDP is infeasible, our method computes an upper bound quantifying the degree to which the SDP is infeasible.
Additionally, it is important to note that our solution to Question 2 will always yield a positive semidefinite matrix and the precision parameter controls the agreement between the approximate solution matrix and the provided entries of $M$.

\section{Semidefinite programming formulation}
We begin with the notation necessary to define the semidefinite programming problem.
Let $\SS ^n_+ := \{X \in \SS^n : X \succeq 0\}$ be the positive semidefinite cone.
Let $\langle \cdot, \cdot \rangle$ denote the standard inner product for vectors and the Frobenius inner product for matrices.
Let $E_{ij} \in \RR^{n \times n}$ be the matrix with a 1 in the $ij^{\textrm{th}}$ position and zeros elsewhere.
For any matrix $M \in \SS^n$, let $\lambda_{\max}(M)$ denote the maximum eigenvalue of $M$ and let $\tr(M)$ denote the trace of $M$.
Let $\cA: \SS^n \to \RR^m$ be a given linear operator and $\cA\adj : \RR^m \to \SS^n$ be its adjoint, i.e. $\langle \cA X, y \rangle = \langle X, \cA\adj y \rangle$ for any $X \in \SS^n$ and $y \in \RR^m$, which take the following form,
\begin{equation*}
   \cA X = \begin{bmatrix} 
   \langle A_1, X \rangle \\
   \vdots \\
   \langle A_m, X \rangle \\
   \end{bmatrix}
    \quad \textrm{and} \quad 
    \cA\adj  y = \sum_{i=1}^m y_i A_i,
\end{equation*}

\noindent For both of the above questions, we can construct the following primal~\eqref{eq:primal_prob} and dual~\eqref{eq:dual_prob} semidefinite programs:
\vspace*{-0.5cm}
\begin{multicols}{2}
\begin{equation*}
    \begin{aligned}
        &\maximize_{X \in \SS^n_+} \quad 0 \\
        \vspace{-20em}
        &\st \;\;\; \cA X = b \\
    \end{aligned}\tag{P}\label{eq:primal_prob}
\end{equation*}
\vfill
  \columnbreak
\begin{equation*}
    \begin{aligned}
        &\minimize_{y \in \RR^m} \quad\langle b, y \rangle \\
        &\st \;\;\; \cA\adj  y \succeq 0
    \end{aligned}\tag{D}\label{eq:dual_prob}
\end{equation*}
\end{multicols}
\noindent where $A_\ell = E_{ij} + E_{ji}$ and $b_\ell = 2 M_{ij}$ for the $\ell^\textrm{th}$ given entry in the $ij^\textrm{th}$ position of $M$. 
Let $\cX_\star$ and $\cY_\star$ denote the solution sets of the~\eqref{eq:primal_prob} and~\eqref{eq:dual_prob}, respectively.
Question 1 is asking whether or not~\eqref{eq:primal_prob} is feasible (i.e. whether $\cX_\star = \emptyset$ or not) and Question 2 is asking for a solution to the primal problem $X_\star \in \cX_\star$.
This formulation allows us to leverage existing fast and scalable SDP optimization algorithms to obtain efficient solutions to the above matrix completion questions.

\section{Spectral bundle method}

We follow the notation and description of the spectral bundle method (SpecBM) provided in~\cite{angell2023fast}.
SpecBM solves the following equivalent penalized formulation of the original dual problem instead of solving either the primal or dual problem directly
\begin{equation}
    \minimize_{y \in \RR^m} f(y) := \alpha \max\{\lambda_\textrm{max}(-\cA\adj y), 0\} + \langle b, y \rangle,\tag{pen-D}\label{eq:pen_dual}
\end{equation}
where $\alpha > \tr(X_\star)$ is the penalty parameter for any $X_\star \in \cX_\star$.
The optimization problem~\eqref{eq:pen_dual} is equivalent to~\eqref{eq:dual_prob} in the sense that they have the same solution set and optimal objective value.
The entire diagonal of $M$ is given in both questions, so the trace of all primal solutions is easily computable.
SpecBM optimizes~\eqref{eq:pen_dual} using a proximal bundle method~\cite{rockafellar1976monotone,mifflin1977algorithm,feltenmark2000dual,kiwiel2000efficiency}.  

\subsection{Proximal bundle method}
Proximal bundle methods are a lesser-known class of optimization algorithms for solving unconstrained convex minimization problems with tremendous upside over other first-order methods~\cite{diaz2023optimal}.
In this case, we will use the proximal bundle method to optimize $f(y)$, the~\eqref{eq:pen_dual} objective.
At each iteration, the proximal bundle method attempts to mimic the proximal point algorithm, proposing an update to the current iterate $y_t$ by applying a proximal step to an approximation of the objective function $\hat{f}_t$, referred to as the \emph{model}:
\begin{equation}
\label{eq:proposal_step}
\tilde{y}_{t+1} \gets \argmin_{y \in \RR^m} \hat{f}_t(y) + \frac{\rho}{2} \| y - y_t \|^2
\end{equation}
where $\rho > 0$.
The next iterate $y_{t+1}$ is set equal to $\tilde{y}_{t+1}$ only when the decrease in the objective value is at least a predetermined fraction of the decrease predicted by the model $\hat{f}_t$, i.e.
\begin{equation}
\label{eq:descent_cond}
\beta (f(y_t) - \hat{f}_t(\tilde{y}_{t+1})) \leq f(y_t) - f(\tilde{y}_{t+1})
\end{equation}
for some fixed $\beta \in (0, 1)$. The iterations where~\eqref{eq:descent_cond} is satisfied (and thus, $y_{t+1} \gets \tilde{y}_{t+1}$) are referred to as \emph{descent steps}. Otherwise, the algorithm takes a \emph{null step} and sets $y_{t+1} \gets y_{t}$. 
Regardless of whether~\eqref{eq:descent_cond} is satisfied or not, $\tilde{y}_{t+1}$ is used to construct the next model $\hat{f}_{t+1}$.

\paragraph{Model Requirements.} The model $\hat{f}_t$ can take many forms, but is usually constructed using subgradients of $f$ at past and current iterates. Let $\partial f(y) := \{g : f(y') \geq f(y) + \langle g, y' - y \rangle, \,\forall y' \in \RR^m\}$ denote the subdifferential of $f$ at $y$ (i.e.\!\! the set of subgradients of $f$ evaluated at a point $y$). Following~\cite{diaz2023optimal}, we require the next model $\hat{f}_{t+1}$ to satisfy some mild assumptions:
\begin{enumerate}
    \item \textit{\textbf{Minorant.}}
    \begin{equation}
    \label{eq:minorant_cond}
        \hat{f}_{t+1}(y) \leq f(y) \quad \textrm{ for all } y \in \RR^m
    \end{equation}
    \item \textit{\textbf{Objective subgradient lower bound.}} For any $g_{t+1} \in \partial f(\tilde{y}_{t+1})$, 
    \begin{equation}
        \label{eq:obj_subgrad_cond}
        \hat{f}_{t+1}(y) \geq f(\tilde{y}_{t+1}) + \langle g_{t+1}, y - \tilde{y}_{t+1}\rangle \quad \textrm{ for all } y \in \RR^m
    \end{equation}
    \item \textit{\textbf{Model subgradient lower bound.}} The first order optimality conditions for~\eqref{eq:proposal_step} gives the subgradient $s_{t+1} := \rho(y_t - \tilde{y}_{t+1}) \in \partial \hat{f}_{t} (\tilde{y}_{t+1})$. After a null step $t$, 
    \begin{equation}
        \label{eq:model_subgrad_cond}
        \hat{f}_{t+1}(y) \geq \hat{f}_{t}(\tilde{y}_{t+1}) + \langle s_{t+1}, y - \tilde{y}_{t+1}\rangle \quad \textrm{ for all } y \in \RR^m
    \end{equation}
\end{enumerate}
The first two conditions are almost inherent, serving to guarantee that a new model integrates first-order information from the objective at $\tilde{y}_{t+1}$. The third condition is workable and, in essence, demands the new model to preserve a degree of the approximation accuracy exhibited by the preceding model.
\citet{angell2023fast} show how to construct, update, and optimize $\hat{f}_t$ efficiently in the specific case of SpecBM. 

\section{Existence of positive semidefinite completions}
In this section, we will present our solution to Question 1. 
First, observe the following facts about this special case SDP related to the optimal objective value and solution set of~\eqref{eq:pen_dual} and feasibility of~\eqref{eq:primal_prob}:
\begin{fact}
    $f(0) = 0$, and thus, the optimal penalized dual objective value $f(y_\star) \leq 0$ for any $y_\star \in \cY_\star$.
\end{fact}
\begin{fact}
    \label{fact:infeas_certificate}
    If there exists $y \in \RR^d$ such that $f(y) < 0$, then~\eqref{eq:primal_prob} is infeasible.
\end{fact}
\begin{fact}
    If~\eqref{eq:primal_prob} is feasible, then $0 \in \cY_\star$.
\end{fact}
\noindent Moreover, if we initialize $y_0 = 0$, then the spectral bundle method will take a descent step only when~\eqref{eq:primal_prob} is infeasible.
In fact, we can determine infeasibility without taking a descent step by Fact~\ref{fact:infeas_certificate} --- i.e. in the case that~\eqref{eq:pen_dual} evaluated at some candidate iterate $f(\tilde{y}_{t+1}) < 0$ even if~\eqref{eq:descent_cond} is not satisfied.

We will now show that SpecBM can be used to answer the decision problem in Question 1 up to some precision $\varepsilon$ (specifically, whether or not $f(y_\star) \geq - \varepsilon$) in $O(1 / \varepsilon)$ iterations.
We will later analyze the per-iteration complexity to obtain the full time complexity. 

We start by defining two quantities used in our analysis.
Let the \emph{proximal gap} at iteration $t$ be defined as follows
\begin{equation}
    \Delta_t := f(y_t) - \left( f(\bar{y}_{t+1}) + \frac{\rho}{2} \|\bar{y}_{t+1} - y_t \|^2\right),
\end{equation}
where $\bar{y}_{t+1} := \argmin_{y \in \RR^m} \{ f(y) + \frac{\rho}{2} \| y - y_t\|^2 \}$. 
Let the \emph{proximal subproblem gap} at iteration $t$ be defined as follows
\begin{equation}
    \widetilde{\Delta}_t := f(y_t) - \left( \hat{f}_t(\tilde{y}_{t+1}) + \frac{\rho}{2} \|\tilde{y}_{t+1} - y_t \|^2\right).
\end{equation}
Observe that $\widetilde{\Delta}_t \geq \Delta_t$ by the minorant requirement~\eqref{eq:minorant_cond}.
In addition, SpecBM efficiently computes $\widetilde{\Delta}_t$ in each iteration, but $\Delta_t$ cannot be efficiently computed (computing $\bar{y}_{t+1}$ requires a full eigendecomposition requiring $O(n^3)$ time).
If~\eqref{eq:primal_prob} is feasible and we initialize $y_0 = 0$, then $\Delta_t = 0$ and $y_t = y_0 = 0$ for all time steps $t$ since no descent steps will ever be taken.
Hence, in a trivial manner $\Delta_t \geq O(f(y_t) - f(y_\star)) = 0.$
Otherwise, suppose that~\eqref{eq:primal_prob} is infeasible, and thus, $f(y_\star) < 0$. In this case, we will utilize standard proximal gap analysis given by~\cite[Lemma 7.12]{ruszczynski2006nonlinear} and amended by~\cite[Lemma 5.3]{diaz2023optimal}, which is reiterated in the following lemma.
\begin{lemma}
    \label{lem:prox_gap_bound}
    Fix a minimizer $y_\star \in \cY_\star$ of $f$ and let $y_t \in \RR^m \setminus \{y_\star\}$. Then, the proximal gap is lower bounded by
    \begin{equation}
        \Delta_t \geq \begin{cases}
            \dfrac{1}{2\rho} \left(\dfrac{f(y_t) - f(y_\star)}{\| y_t - y_\star \|}\right)^2 & \textrm{if } f(y_t) - f(y_\star) \leq \rho \| y_t - y_\star \|^2, \vspace{0.5em}\\ 
            f(y_t) - f(y_\star) - \dfrac{\rho}{2} \| y_t - y_\star \|^2 & \textrm{otherwise.}
        \end{cases}
    \end{equation}
\end{lemma}
\noindent We can assume without loss of generality that $y_t = y_0 = 0$, and thus, we can rewrite the smoothing parameter as follows
\begin{equation}
    \rho = \lambda \frac{f(y_t) - f(y_\star)}{\|y_t - y_\star\|^2}  = \lambda \frac{-f(y_\star)}{\|y_\star\|^2},
\end{equation}
where $\lambda > 0$ is constant for all null step iterations and $\|y_\star\| \ne 0$ by the assumption that $f(y_\star) < 0$.
Using this realization, we can simplify the bound from Lemma~\ref{lem:prox_gap_bound} as follows
\begin{equation}
        \Delta_t \geq \begin{cases}
            \dfrac{1}{2\lambda} (f(y_t) - f(y_\star)) & \textrm{if } \lambda > 1, \vspace{0.75em}\\
            \dfrac{1}{2} (f(y_t) - f(y_\star)) & \textrm{otherwise}.
        \end{cases}
\end{equation}
Hence, $\widetilde{\Delta}_t \geq \Delta_t \geq O(f(y_t) - f(y_\star))$ independent of whether~\eqref{eq:primal_prob} is feasible or not, and thus, it suffices to show $ \widetilde{\Delta}_t \leq \varepsilon$ for $t \leq O(1 / \varepsilon)$ iterations (specifically, null steps) to obtain the desired result.
The core of this sufficient result is given by the following lemma.
\begin{lemma}
For every null step $t$, $\widetilde{\Delta}_t$ decreases according to the following recurrence
\begin{equation}
    \widetilde{\Delta}_{t+1} \leq \widetilde{\Delta}_t - \frac{(1-\beta)^2\rho}{8G^2} \,\widetilde{\Delta}_t^2,
\end{equation}
where $G = \sup_{t \geq 0} \{\|g_{t+1}\| : g_{t+1} \in \partial f(\tilde{y}_{t+1})\}$.
\end{lemma}
\noindent This result is shown in the proof of~\cite[Lemma 5.2]{diaz2023optimal}.
Using \cite[Lemma A.1]{diaz2023optimal}, we can conclude that this recurrence gives $\widetilde{\Delta}_t \leq \varepsilon$ after $t \leq O(1 / \varepsilon)$ null steps.
We note that it may be possible to improve this rate under certain assumptions and SpecBM parameter settings as suggested by the results of~\citet{ding2023revisiting}.

\subsection{Per-iteration analysis}
For ease of per-iteration analysis, we consider simplest case of SpecBM~\cite{angell2023fast} when the parameter setting is $k_c = 1$ and $k_p = 0$. This parameter setting greatly simplifies the steps in each iteration. Increasing these parameters sacrifices per-iteration time complexity and memory usage for improved convergence. 
Under this simplest parameter setting, we analyze the per-iteration time complexity to obtain the full computational complexity.

First, we note that SpecBM does not require storing the primal iterates $X_t$ explicitly.
In fact, for this special SDP instance, we only need to store $\tr(X_t)$ and $\cA X_t \in \RR^m$.
Both of these quantities can be maintained efficiently with low-rank updates due to the linearity of both operations.

In each iteration, SpecBM requires computing a maximum eigenvector (and associated eigenvalue) of $-\cA\tilde{y}_{t+1}$, a sparse matrix with $m$ entries. 
For the sake of analysis, we can compute a maximum eigenvector using a randomized Lanczos method~\cite{lanczos1950iteration}.
In practice, the randomized Lanczos method has known numerical stability issues and we would advise using the thick-restart Lanczos method~\cite{hernandez2007krylov, wu1999thick} as is implemented in~\cite{angell2023fast}.
~\citet[Theorem 4.2(a)]{kuczynski1992estimating} showed that the randomized Lanczos method can compute an $\varepsilon$-approximate maximum eigenvector in $\tilde{O}(1 / \sqrt{\varepsilon})$ iterations with high probability.
We also note that the randomized Lanczos method will not run for more than $n$ iterations (the maximum possible number of $n$-dimensional orthonormal vectors is $n$).
Note that we are assuming the inner approximation error is the same as the outer approximation error, which is unlikely to be true in practice, but does not effect the spirit or merit of our results.
Each iteration of the randomized Lanczos method is dominated by a sparse matrix-vector multiplication which takes $O(m)$ time.
Hence, an $\varepsilon$-approximate maximum eigenvector can be computed in $\tilde{O}(\min\{nm, m / \sqrt{\varepsilon}\})$ time.

Under the parameter setting $k_c=1$ and $k_p=0$, all of the other per-iteration operations of SpecBM can be computed in $\tilde{O}(m)$ operations.
Hence, each iteration takes $\tilde{O}(\min\{nm, m / \sqrt{\varepsilon}\})$ and yielding the full complexity of $\tilde{O}(\min\{nm/\varepsilon, m / \varepsilon^{3/2}\})$.

\section{Fast approximate squaring}

In this section, we will present our solutions to Question 2 and we will assume without loss of generality that~\eqref{eq:primal_prob} is feasible.
We present two solutions, one without matrix sketching and one with matrix sketching.
In our solutions, we define an $\varepsilon$-approximate solution to be $\| \cA X_t - b\| \leq \varepsilon$.
Additionally, note that all of the primal iterates $X_t$ produced by SpecBM are positive semidefinite.

We will first show that SpecBM produces primal iterates such that $\|\cA X_t - b\| \leq \varepsilon$ in $O(1/\varepsilon^2)$ iterations.
First, using the closed-form solution for the next candidate iterate $\tilde{y}_{t+1}$ given in~\cite{angell2023fast}, it can be seen that
\begin{equation}
    \tilde{y}_{t+1} = \frac{1}{\rho}(\cA X_{t+1} - b).
\end{equation}
Using this closed-form solution and the assumption that $y_t = 0$, observe that
\begin{equation}
    \begin{aligned}
        \widetilde{\Delta}_t &= f(y_t) -\left(\hat{f}_t(\tilde{y}_{t+1}) + \frac{\rho}{2} \|\tilde{y}_{t+1} - y_t \|^2 \right) \\
        &= -\left(\langle -\cA\adj\tilde{y}_{t+1}, X_{t+1}\rangle + \langle b, \tilde{y}_{t+1} \rangle + \frac{\rho}{2} \left\| \tilde{y}_{t+1} \right\|^2\right) \\
        &= -\left(-\frac{1}{\rho}\langle \cA X_{t+1} - b, \cA X_{t+1} - b \rangle + \frac{\rho}{2} \left\| \frac{1}{\rho} (\cA X_{t+1} - b)\right\|^2\right) \\
        &= \frac{1}{2\rho} \| \cA X_{t+1} - b\|^2.
    \end{aligned}
\end{equation}
We already showed in the previous section that $\widetilde{\Delta}_t \leq \varepsilon$ after $t \leq O(1/\varepsilon)$ iterations, so we can conclude that $\|\cA X_t - b\| \leq \varepsilon$ after $t \leq O(1 / \varepsilon^2)$ iterations.

Since SpecBM does not require explicitly storing the primal iterates $X_t$, we have two options for keeping track of $X_t$ via low-rank updates (under this simplified parameter setting they are rank-one updates).
We can update $X_t$ explicitly in $O(n^2)$ time per-iteration.
We can also save time and space using a Nystr\"om sketch~\cite{tropp2017fixed, gittens2013topics, halko2011finding, li2017algorithm}.
Since the SDP has $m$ constraints and~\eqref{eq:primal_prob} is feasible, there exists a solution to~\eqref{eq:primal_prob} with rank $r \leq \sqrt{2(m + 1)}$~\cite{pataki1998rank,barvinok2002course}.
This means we can use an $O(\sqrt{m})$-dimensional Nystr\"om sketch to store and efficiently update a random projection of $X_t$ in $O(n\sqrt{m})$ time and space (see~\cite{yurtsever2021scalable, angell2023fast} for how to reconstruct a factorized approximation of $X_t$).
This yields the full complexity results of $\tilde{O}(n^2 / \varepsilon^2 + \min\{ nm / \varepsilon^2, m / \varepsilon^{5/2}\})$ and $\tilde{O}(n\sqrt{m}/\varepsilon^2 + \min\{ nm / \varepsilon^2, m / \varepsilon^{5/2}\})$ for without and with sketching, respectively.

\bibliographystyle{plainnat}
\bibliography{main}

\pagebreak

\end{document}